\documentclass[11pt]{article}

\usepackage{multirow,array}
\usepackage{xfrac}
\usepackage{svg}
\usepackage[utf8]{inputenc}
\usepackage[english]{babel}
\usepackage{enumitem}
\usepackage{booktabs}       % professional-quality tables
\usepackage{amsfonts}       % blackboard math symbols
\usepackage{nicefrac}       % compact symbols for 1/2, etc.
\usepackage{microtype}      % microtypography
\usepackage[cmex10]{amsmath}
\usepackage{amsthm}
\usepackage{amssymb}
\usepackage{lipsum}

\usepackage{balance} 

\usepackage[linesnumbered,algoruled,boxed,lined]{algorithm2e}

\begin{document}

\title{Egocentric Bias and Doubt in Cognitive Agents}
\author{Nanda Kishore Sreenivas \hspace{2cm} Shrisha Rao}
\date{}
\maketitle

\begin{abstract}  % put your abstract here!

Modeling social interactions based on individual behavior has always
been an area of interest, but prior literature generally presumes
rational behavior.  Thus, such models may miss out on capturing the
effects of biases humans are susceptible to.  This work presents a
method to model egocentric bias, the real-life tendency to emphasize
one's own opinion heavily when presented with multiple opinions. We
use a symmetric distribution centered at an agent's own opinion, as
opposed to the Bounded Confidence (BC) model used in prior work.  We
consider a game of iterated interactions where an agent cooperates
based on its opinion about an opponent.  Our model also includes the
concept of domain-based self-doubt, which varies as the interaction
succeeds or not.  An increase in doubt makes an agent reduce its
egocentricity in subsequent interactions, thus enabling the agent to
learn reactively.  The agent system is modeled with factions not
having a single leader, to overcome some of the issues associated with
leader-follower factions.  We find that agents belonging to factions
perform better than individual agents.  We observe that an
intermediate level of egocentricity helps the agent perform at its
best, which concurs with conventional wisdom that neither
overconfidence nor low self-esteem brings benefits.

\end{abstract}

\noindent{\bf Keywords}: egocentric bias; cognitive psychology; doubt;
factions; Continuous Prisoner's Dilemma; opinion aggregation

\section{Introduction}

Decision making has been a long-studied topic in the domain of social
agent-based systems, but most earlier models were rudimentary and
assumed rational behavior~\cite{Schelling,sugarscape}. Decision making
is strongly driven by the opinion that an agent holds; this opinion is
shaped over time by its own initial perceptions \cite{perception}, its
view of the world \cite{eiser}, information it receives over various
channels, and its own memory \cite{mem}. This process of opinion
formation is fairly complex even under the assumption that Bayesian
reasoning applies to decision making. Years of research in psychology
has shown that humans and even animals~\cite{rats,Burman} are
susceptible to a wide plethora of cognitive
biases~\cite{cblist}. Despite the immense difficulties in the
understanding and description of opinion dynamics, it continues to be
an area of immense interest because of the profound impact of
individual and societal decisions in our everyday lives.

In this work, we model agents with \emph{egocentric bias} and focus on
agents' opinion formation based on its perception, memory, and
opinions from other agents.  Egocentric bias may be described as the
tendency to rely too heavily on one's own perspective. The bias has
been claimed to be ubiquitous~\cite{nic} and
ineradicable~\cite{Krueger}.  Egocentric bias is commonly thought of
as an umbrella term, and covers various cognitive biases, including
the anchoring bias~\cite{anchoring2,anchoring1,anchoring0} and the
``false consensus effect''~\cite{anchoring5}.  Recent research seems
to suggest that such bias is a consequence of limited cognitive
ability~\cite{anchoring3} and the neural network structure of the
brain~\cite{anchoring4}.

The initial approaches to model opinion dynamics were heavily inspired
by statistical physics and the concept of atomic spin states.  They
were thus detached from real life and allowed only two levels of
opinions~\cite{cox,sznajd}. There was also the social impact
model~\cite{Nowak96,NowakLatane90} and its further variations with a
strong leader~\cite{nowakleader}.

Later models, such as the ones proposed by
Krause~\cite{Krause_adiscrete} and Hegselmann~\cite{Opn1} considered
continuous values for opinions and introduced the \emph{Bounded
  Confidence} (BC) model which incorporated the \emph{confirmation
  bias}. The BC model and the relative agreement model by Deffuant
\emph{et al.}~\cite{Deffuant2002} inspired by the former, have
remained in favor until now~\cite{allahverdyan,li2017}. Confirmation
bias has also been modeled in other contexts, such as
trust~\cite{trustCB} and conversational agents~\cite{bouchet,hayashi}.

Historically, models concentrating on opinion dynamics have revolved
around consensus formation.  However, opinions are not formed at an
individual or societal level without any consequence. Rather, these
opinions lead to decisions and these decisions have a cost, a result,
and an outcome.  In reality, humans as well as animals learn from
outcomes and there are subtle changes introduced in this process of
opinion formation during subsequent interactions.

The assignment of weights to all opinions including one's own is a
major issue in opinion formation. In the BC model, the weights are
taken as a uniform distribution within the interval, and opinions
outside of this are rejected.  The problem with this model is that it
is too rigid. To introduce some level of flexibility into our model,
we consider the assignment of weights as per a symmetric distribution
centered around the agent's perspective with its flatness/spread
varied according to that agent's level of egocentricity.

We consider a game of iterated interactions where an agent, say $A$,
is paired with some random agent, say $B$, in one such iteration. Each
of these interactions is a Continuous Prisoner's Dilemma (CPD)
\cite{CPD}, which allows an agent to cooperate at various levels
bounded by $[0,1]$.  Here, \emph{opinions} are based upon an agent's
knowledge about the opponent's level of cooperation in prior
interactions and thus lies between 0 and 1. Thus, $A$ has its own
opinion of $B$ and it also takes opinions of $B$ from other
sources. $A$ aggregates all the opinions and cooperates at that level,
and it then decides the outcome of this interaction based on $B$'s
level of cooperation.

Our model also captures an agent's reaction to this outcome. When an
agent succeeds, there is a rise in self-esteem and this is reflected
in a higher egocentricity in subsequent interactions. We model
reaction to failure as a loss of self-esteem i.e., a rise in
\emph{self-doubt} on this domain \cite{esteem}.  This domain-based
self-doubt is a key aspect of this model as it helps an agent to learn
reactively.

In our model, agents can belong to \emph{factions} as well. While most
works have modeled factions as a leader-followers structure
\cite{fac,Atwater1999}, we model a faction with a central memory, that
holds the faction's view on all agents in the system. The faction's
view is an unbiased aggregate of individual opinions of its
members. To sum up, an agent can have up to three different levels of
information---its own opinion, opinions from friends, and the
faction's view.

Through simulation, we find results about optimum level of
egocentricity and the effect of faction sizes. Varying the levels of
egocentricity among agents, it is observed that agents with an
intermediate level perform much better than agents with either low or
high levels of egocentricity. This is in strong agreement with
conventional wisdom that neither overconfidence nor low
self-confidence brings optimum results. Agents in larger factions are
observed to perform better, and results indicate a linear
proportionality between value and faction size as suggested by
Sarnoff's Law. Also, to understand the effects of other attributes of
the system, we vary the number of interactions, the proportion of
different agents, the types of agents, etc.

\section{Related Work}

In his work on opinion dynamics~\cite{Sobkowicz}, Sobkowicz writes:

\begin{quote}
  ``Despite the undoubted advances, the sociophysical models of the
  individual behaviour are still rather crude. Most of the
  sociophysical agents and descriptions of their individual behaviour
  are too simplistic, too much `spin-like', and thus unable to capture
  the intricacies of our behaviours.''
\end{quote}

Our work thus focuses on three key aspects---egocentricity,
self-doubt, and the concept of factions. In this section, we review
the existing work in these domains. 

\noindent\emph{Egocentric Bias}

Egocentric bias is the tendency to rely too heavily on one's own
perspective and/or to have a higher opinion of oneself than
others. Ralph Barton Perry~ \cite{Ralph} coined the term
\emph{egocentric predicament} and described it as the problem of not
being able to view reality outside of our own
perceptions. Greenwald~\cite{Greenwald} described it as a phenomenon
in which people skew their beliefs in agreement with their perceptions
or what they recall from their memory.  We are susceptible to this
bias because information is better encoded when an agent produces
information actively by being a participant in the interaction.

Research suggests that this skewed view of reality is a virtually
universal trait and that it affects each person's life far more
significantly than had been realized~\cite{nic}. It has also been
shown to be pervasive among people and groups in various contexts such
as relationships, team sports, etc.~\cite{Ross}. It is closely
connected to important traits such as self-esteem and
confidence~\cite{kimberly}. A high degree of egocentric bias hinders
the ability to empathize with others' perspectives, and it has been
shown that egocentricity tends to be lower in depressed
individuals~\cite{NY}. Egocentric bias also plays a key factor in a
person's perception of fairness: people tend to believe that
situations that favor them are fair whereas a similar favor to others
is unjust~\cite{Feng2018,Greenberg}. Perceived fairness is a crucial
element in several resource allocation problems. Most importantly, it
has been shown to be \emph{ineradicable} even after standard debiasing
strategies such as feedback and education~\cite{Krueger}.

Prior work has been done to model confirmation bias, but the most used
model has been the Bounded Confidence (BC) model. The BC model was
first introduced by Krause in 2000~\cite{Krause_adiscrete}. Later,
Deffuant \emph{et al.}~\cite{Deffuant2002} proposed a relative
agreement model (RA) which extended the BC model. In the BC model, an
agent considers only those opinions that are sufficiently close to its
own, and shuns any opinion outside the confidence threshold. This
model has been used to model confirmation bias in many
papers~\cite{Weibush,Deffuant2000,Opn1,Sobkowicz,delVic}.

\emph{Self-doubt} 

There can be multiple responses to a perceived failure---lowering of
one's aspiration, loss of self-esteem manifested as an increase in
doubt, or even leaving the activity altogether \cite{lewin}.

The term \emph{self-esteem} has been used in three ways---global
self-esteem, state self-esteem and domain specific self-esteem
\cite{3faces}. We are primarily concerned with an agent's
domain-specific self-esteem in this paper, which is a measure of one's
perceived confidence pertaining to a single domain.  Our work models
the self-doubt which is a counterpart of this. Self-doubt is defined
as ``the state of doubting oneself, a subjective sense of instability
in opinions''~\cite{selfdoubt}.

\emph{Factions}

Factions have been broadly considered to be specific sets of
agents. However, a faction has been modeled in different ways. Some
factions have been modeled as a leader-follower group, where the
leader determines the group dynamics~\cite{fac}. Even if the group
does not have an assigned leader to start with, it has been suggested
that an agent with high cognitive capacity eventually emerges as a
leader~\cite{Atwater1999}.  Such a leader eventually impacts the
performance of the entire group. Factions can also be modeled as a
selfish herd, where each agent is a member for its own
gain~\cite{herd1}. However, this structure does not have a single
leader and such models have proved useful in modeling certain group
behaviors~\cite{herd2,herd3}.

\section{Egocentric Interactions}

We consider a system of agents playing a game of iterated
interactions. In each iteration, an agent $A$ is paired with some
agent $B$ randomly. Since the model is based on the Continuous
Prisoner's Dilemma (CPD)~\cite{CPD}, an agent can cooperate at any
level between 0 and 1, with 0 corresponding to defection and 1 to
complete cooperation. $C_B(t)$ denotes $B$'s level of cooperation with
$A$ in interaction $t$, and this value lies between 0 and 1.

The opinion of $A$ about $B$ at the next interaction $t+1$, denoted by
$\eta_A(B,t+1)$ is based on $A$'s previous $\omega$ previous
experiences with $B$, where $\omega$ is the memory size:
\begin{equation}
    \eta_A(B,t) = \frac{C_B(t-1) + C_B(t-2) + \hdots + C_B(t-\omega)}{\omega}
    \label{eq:opeq}
\end{equation}

$A$ has its own opinion of $B$ and also collects opinions of $B$ from
its friends (described in Section \ref{social_structure}) as well. It
aggregates these opinions according to its egocentricity, and this
aggregate is used as its level of cooperation in the next interaction
with $B$.  If $B$ cooperates at a satisfactory level, $A$ decreases
its doubt on $B$ and thus is more egocentric in the next interaction
with $B$. These concepts and corresponding formulations are outlined
in the following subsections.

\subsection{Egocentricity}\label{egocentricity}

As discussed in the previous section, current models of egocentricity
consider a Bounded Confidence (BC) model and all opinions within this
interval get the same weight
\cite{Krause_adiscrete,Deffuant2002,Sobkowicz}. This uniform
distribution of weights across the confidence interval is not an
accurate depiction because such a model would assign the same weight
to one's own opinion and an opinion on the fringes of the
interval. Also, an opinion that is outside the interval by a mere
fraction is to be completely rejected, which is too rigid. This raises
the need for some flexibility, and hence we use a Gaussian (normal)
distribution to calculate the weights.  The use of a symmetric
distribution to model the agent's judgments with egocentric bias is a
manifestation of the anchoring-and-adjustment heuristic, which has a
neurological basis~\cite{Tamir2010} and is well known in studies of
the anchoring bias~\cite{anchoring0,anchoring2}.  The same type of
distribution is seen in the context of anchoring bias in the work of
Lieder \emph{et al.}~\cite{anchoring3}.  The mean of the curve is the
agent $A$'s opinion of the other agent $B$, as the mean gets the
highest weight in this distribution.
\begin{figure}[!htbp]
    \setlength{\abovecaptionskip}{0pt plus 1pt minus 1pt} 
    \centering
    \includegraphics[width=0.8\textwidth,height=0.6\textheight,keepaspectratio]{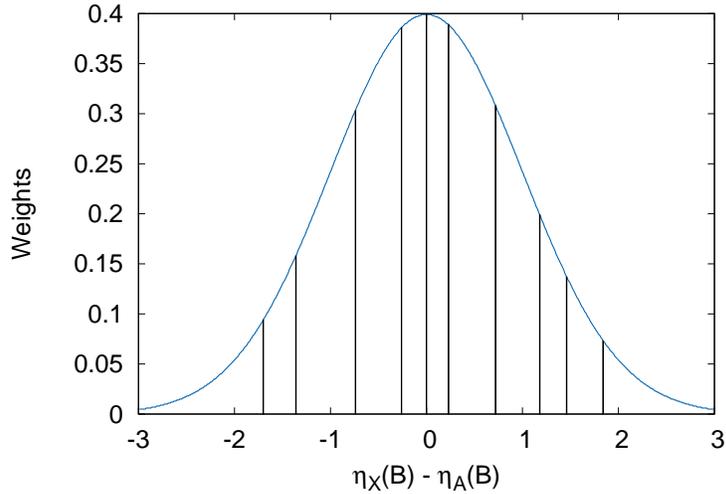}
    \caption{Weights assigned for a range of opinions}
    \label{fig:ch}
\end{figure} 
%\vspace-0.1 in}

Each agent has a base egocentricity $E_0$, which is a trait of the
agent and remains constant.  The egocentricity of an agent $A$ with
respect to $B$ is obtained by factoring the doubt of $A$ about $B$ on
$A$'s base egocentricity. This egocentricity is manifested as the
spread of the Gaussian curve, $\sigma$. The base spread $\sigma_0$ is
inversely proportional to the base egocentricity:
\begin{equation}
    \sigma_0 = \frac{1}{E_0}
    \label{eq:inverseeq}
\end{equation}

The higher the egocentricity of an agent, the lower is the spread of
the curve, thus assigning a relatively high weight to its own opinion.
The lower the egocentricity of the agent, the flatter is the curve,
thus assigning roughly the same weight to all opinions including its
own.  However, this spread depends on agent $B$ and is an adjusted
value of the base spread, $\sigma_0$.

\subsection{Self-Doubt}\label{doubt}

Literature in psychology suggests several strategies as responses to
failure---quitting, readjusting levels of aspiration, and increasing
self-doubt~\cite{lewin}.  In this work, we model an increase in
domain-specific self-doubt as the response to failure, since
self-doubt has been claimed to be useful~\cite{woodman}.  In response
to a successful interaction, an agent gains in self-confidence, i.e.,
the self-doubt is decreased. To classify an interaction as success or
failure, $A$ has a threshold of \emph{satisfaction} $\lambda$; $B$'s
level of cooperation has to be higher than $\lambda$ for $A$ to deem
the interaction successful.

Self-doubt is used by agent $A$ as a multiplicative factor on the base
spread to obtain the relevant spread for aggregating opinions about
agent $B$.  Since doubt is a multiplicative factor, doubt about all
agents is initialized as 1, and agent $A$ uses $\sigma_0$ as the
spread initially.  With subsequent interactions, doubt varies as
described, and this ensures a constantly changing level of
egocentricity, based on outcomes of previous interactions. Since the
spread has to be positive at all times, doubt is lower-bounded by
0. Theoretically there is no fixed upper bound for doubt, but beyond a
certain value, the curve gets flat enough to a point where all
opinions effectively get the same weight. As doubt tends to zero, the
agent is completely certain of its opinion and rejects all other
opinions.  Doubt is subjective, and the doubt of $A$ about $B$ is
denoted by $\mathcal{D}_A(B)$.  It is updated after each interaction
between $A$ and $B$, being incremented or decremented by a constant
$c$ depending on whether $A$ is satisfied or dissatisfied with the
interaction.

The spread of the Gaussian curve used for assigning weights to
opinions about $B$ is calculated considering the agent-specific doubt
and its base spread, $\sigma_0$:
\begin{equation}
    \sigma = \sigma_0\mathcal{D}_A(B)
    \label{eq:sigmaeq}
\end{equation}

This $\sigma$ defines the spread of the normal distribution used by
$A$ to assign weights to different opinions on $B$.

\subsection{Social Structures} \label{social_structure}

Though leader-follower models are seen in certain contexts, such as in
the context of modeling countries~\cite{Silverman2007},
oligopolies~\cite{oligopolies}, and insurgents~\cite{Silverman2008},
there is evidence that other models with no explicit leader-follower
structure are appropriate to understand many societal behaviors.  The
``selfish herd'' model~\cite{herd1} was originally suggested for
animals seeking to avoid predation and other dangers, but it is seen
to explain human social behavior as well~\cite{herd5,herd2}.  Such
models explain economic behavior~\cite{herd4} as well as the evolution
of fashions and cultural changes in society~\cite{herd3}.  Social
media as well as online purchases are also best explained in this
way~\cite{herd6}.  Group decision making in humans does not follow a
strict leader-follower structure even in the presence of
bias~\cite{group1}, and the same is also true when it comes to solving
complex problems by teams~\cite{group2}.  Gill~\cite{group3} gives the
example of Wikipedia as a well-known example of ``collaborative
intelligence'' at work.

Therefore, we model our agent system as being split into factions,
without any single authoritarian faction leader who sets the faction's
view.  Rather, each faction is modeled to be a herd where all members
contribute towards the formation of a \emph{central memory}, which
holds an unbiased aggregate of member opinions about each agent in the
system.

The contribution of all members is assumed to be authentic and
complete.  However, the level of adherence of the faction's view is
different for each agent.  Some agents can be modeled as extreme
loyalists, who suspend their own judgment and simply adhere to the
faction view, while there are others who are individualists and do not
always conform.  We introduce the notion of faction alignment
$\kappa$, which is a measure of an agent's adherence with its faction,
with 0 indicating total nonconformance and 1 complete adherence.

The \emph{friends} of an agent $A$ are a small subset of the agents in
$A$'s faction. The number of friends may be different for each agent
and friendships are defined at random when factions are initialized,
but remain intact thereafter.  A \emph{friendship} is the two-way
connection between two friends.  Based on the seminal work of
Dunbar~\cite{dunbar1} on the number of friends in primate societies, a
recent paper suggests the concept of Dunbar layers~\cite{dunbar2}---an
individual's network is layered according to strength of emotional
ties, with there being four layers in all and the two outermost layers
having 30 and 129 members, which suggests that the average number of
friends is about 25\% of the overall social circle.  As the number of
friends for an individual is variable, as is the total number of
friendships in the faction, we use $\sfrac{z^2}{8}$ as an upper bound
for the number of friendships within the faction, where $z$ is the
faction size.  Friends are the only source of opinions for an agent.
Agents fully cooperate when they interact with a friend.

\subsection{Game Setting} 

The standard Prisoner’s Dilemma (PD) is discrete, so each agent can
choose one of only two possible actions: cooperate or defect.
However, not all interactions can be modeled perfectly by such extreme
behavior.

In the Continuous Prisoner's dilemma (CPD)~\cite{CPD,CPD2}, a player
can choose any level of cooperation between 0 and 1.  We borrow the
concept and the related payoff equations from Verhoeff's work on the
Trader's Dilemma~\cite{CPD}.  Here, a cooperation level of 0 and 1
correspond to the cases of complete defection and complete cooperation
respectively in the PD.

Consider two agents $A$ and $B$ in a CPD, with their cooperation
levels being $a$ and $b$ respectively. The payoff functions are
obtained~\cite{CPD} from the discrete payoff matrix by linear
interpolation:

\begin{equation}
    p_A(a,b)=abC + a\Bar{b}S + \Bar{a}bT + \Bar{a}\Bar{b}D
    \label{eq:payoffeq}
\end{equation}
where $C, T, D, S$ are the payoffs in the standard PD as shown below
%\vspace-0.2in}
\begin{table}[!htbp]
    \setlength{\extrarowheight}{2pt}
    \begin{tabular}{cc|c|c|}
      & \multicolumn{1}{c}{} & \multicolumn{2}{c}{Player $B$}\\
      & \multicolumn{1}{c}{} & \multicolumn{1}{c}{$1$}  & \multicolumn{1}{c}{$0$} \\\cline{3-4}
      \multirow{2}*{Player $A$}  & $1$ & $(C,C)$ & $(S,T)$ \\\cline{3-4}
      & $0$ & $(T,S)$ & $(D,D)$ \\\cline{3-4}
    \end{tabular}
\end{table}

%\vspace-0.1in}

The conditions for choosing the values of these variables are $2C > T
+ S$ and that $T > C > D > S$.  Most work on PD, including Axelrod's
seminal work on evolution of cooperation~\cite{Axel}, uses this set of
values: $\langle C = 3, T = 5, D = 1, S = 0 \rangle$, and we do the
same.

\subsection{Opinion Aggregation}\label{aggr}

There are three phases in each interaction between two agents $A$ and
$B$:

\begin{enumerate}
    \item \emph{Phase 1}: $A$ adjusts its own opinion $\eta_A(B)$ and
      all opinions it has received from its friends
      $\{\eta_{f_1}(B),\eta_{f_2}(B), \ldots \}$, with weights
      represented by vector $W$ to form an intermediate opinion, $O'$.
    \item \emph{Phase 2}: $A$ incorporates $\mathcal{M}_F(B)$, the
      faction's view about $B$, to the intermediate opinion $O'$ using
      its faction alignment $\kappa$ as the weight.
    \item \emph{Updates}: The interaction takes place, payoff $\rho_A$
      is updated, the outcomes classified according to $A$'s
      satisfaction $\lambda$, and doubt $D_A(B)$ is updated.
\end{enumerate}

Consider an agent $A$, which has $m$ friends $ \langle f_1,f_2,\hdots
f_m \rangle$, wishing to form an informed opinion about another agent
$B$, given its own and its friends' opinions of $B$.

\noindent\emph{Phase 1} 

As per the definition of opinion in (\ref{eq:opeq}), the opinions of
$B$ by $A$ and its friends can be structured as a vector $E$, given by
\begin{equation}
    E =  \begin{bmatrix}
            \eta_A(B)\\
            \eta_{f_1}(B)\\
            \eta_{f_2}(B)\\
            \vdots\\
            \eta_{f_m}(B)
            \end{bmatrix}
    \label{eq:eeq}
\end{equation}

The corresponding weights to each opinion are denoted by the vector
$W$ as,
\begin{equation}
    W = \begin{bmatrix}w_A & w_{f_1} & w_{f_2} & \hdots & w_{f_m}\\\end{bmatrix}
    \label{eq:weq}
\end{equation}

Our main problem here is to come up with a $W$ that takes $A$'s
egocentricity into account.  As described in
Section~\ref{egocentricity}, we consider a normal probability
distribution for this purpose.
\begin{equation}
   w_x = \frac{1}{{\sigma \sqrt {2\pi } }} e^{{{ - \left( {\eta_x(B) - \mu } \right)^2 } \mathord{\left/ {\vphantom {{ - \left( {x - \mu } \right)^2 } {2\sigma ^2 }}} \right. \kern-\nulldelimiterspace} {2\sigma ^2 }}}    
   \label{eq:NormalEq}
\end{equation}

\begin{center}
   where $\mu = \eta_A(B) , \, \,
    \sigma = \sigma_0 \times \mathcal{D}_A(B)$ 
\end{center}
So, $O'$, the opinion at the end of Phase 1, is given by
\begin{equation}
   O' = W \cdot E
   \label{eq:doteq}
\end{equation}

This can also be written in an algebraic form as
\begin{equation}
   O' =  w_A\eta_A(B) + \sum_{i=1}^{m} w_{f_i}\eta_{f_i}(B)
\end{equation}

\emph{Phase 2}

Phase 2 of opinion formation focuses on incorporating the faction's
view of agent $B$ into the opinion arrived at in phase 1, $O'$. Let
the faction view on $B$ be denoted by $\mathcal{M}_F(B)$ and let
$\kappa_A$ represent $A$'s level of alignment towards its faction.
Now, the final opinion about $B$ is a $\kappa$-weighted average of
$O'$ and $\mathcal{M}_F(B)$
\begin{equation}
    O = \kappa_A \mathcal{M}_F(B) + (1-\kappa_A) O'
    \label{eq:finaleq}
\end{equation}

\emph{Updates}

The updates phase starts off with updating the payoff $\rho_A$
according to (\ref{eq:payoffeq}).
\begin{equation}
    \rho_A = \rho_A + p_A(a,b)
    \label{eq:payoffupdateeq}
\end{equation}

Based on the outcome of this interaction with $B$ (the level of
cooperation $b$), $A$ updates $\mathcal{D}_A(B)$, its doubt about $B$.
For $A$ to classify its interaction as successful, $b$ has to be
greater than $\lambda$. $\mathcal{D}_A(B)$ is decremented by a
constant $c$ if the interaction is successful, and it is incremented
by $c$ otherwise, as outlined in Section~\ref{doubt}.
\begin{equation}
    \mathcal{D}_A(B)=\left\{\begin{array}{cl}
            \mathcal{D}_A(B)+c,& b < \lambda_A\\
\mathcal{D}_A(B)-c,& b > \lambda_A\end{array}\right.
\label{eq:updateeq}
\end{equation}

Thus, $A$ aggregates opinions about $B$ received from its friends,
taking into account its level of egocentricity and its doubt about
$B$.

\section{Agent Types}\label{types}

An agent pool consisting of three types of agents is considered. The
agents are categorized into different types based on their internal
working and attributes. The system is initially configured with
attributes such as the total number of agents, the proportion of
different types, the number of factions in the system, and the number
of iterations. In each iteration, an agent is randomly paired with one
other agent, then they interact, and finally, each agent updates its
experiences and payoff. We formally define the various attributes of
an agent before delving into the intricacies of each type.

\noindent\emph{Basic Attributes}

All agent types have four basic attributes as described below:
\begin{enumerate}
    \item[$\bullet$] $\alpha$ is a unique identifier for each agent,
      $\alpha \in \{1, \ldots, N\}$, where $N$ is the number of agents
      in the system.
    \item[$\bullet$] $\rho$ is the agent's cumulative payoff, a metric
      to capture the efficiency or performance, $\rho \in \mathbb{Z}^+
      \cup \{0\}$, where $\mathbb{Z}^+$ denotes the set of positive
      integers.
    \item[$\bullet$] $\omega$ is the memory size of an agent, $\omega
      \in \mathbb{Z}^+$.
    \item[$\bullet$] $\mathcal{E}$, the experiences is a
      two-dimensional vector with $N$ rows and $\omega$
      columns. $\mathcal{E}[i][j] = C_i(t-j)$, where $C_X(t)$ denotes
      $X$'s level of cooperation in interaction $t$.
    \begin{center}
    $\mathcal{E} =\begin{bmatrix}
    C_1(t-1) & C_1(t-2) & \dots & C_1(t-\omega) \\
    C_2(t-1) & C_2(t-2) & \dots & C_2(t-\omega) \\
    \hdotsfor{4} \\
    \hdotsfor{4} \\
    C_N(t-1) & C_N(t-2) & \dots & C_N(t-\omega)
    \end{bmatrix}$
    \end{center}
\end{enumerate}

\noindent\emph{Extended attributes}

Apart from the basic attributes, an agent type may also have several
other attributes that enable their functionality and behavior. We
describe them as follows:

\begin{itemize}
    \item $\mathcal{D}$ represents the self-doubt of an agent. It is a
      vector indexed by agent id and reflects the level of uncertainty
      of the agent's own opinion about the corresponding other agent.
      For agent $A$, $\mathcal{D}_A =
      [\mathcal{D}_A(1),\mathcal{D}_A(2),...\mathcal{D}_A(N)]$, $0 <
      \mathcal{D}_A(j) < \infty$.
    \item $\mathcal{F}$ is the set of friends of an agent. Friends of
      an agent $A$ can be described as a subset of its faction,
      $\mathcal{F}_A \subset \Upsilon$ (see below).
    \item $\sigma_0$ represents the base spread of an agent, as
      defined in (\ref{eq:inverseeq}), $\sigma_0 \in \mathbb{R}^+$,
      where $\mathbb{R}^+$ is the set of positive real numbers.
    \item $\lambda$ represents the threshold of satisfaction of an
      agent, as outlined in Section~\ref{aggr}; $0.5 \leq \lambda <
      1$.
    \item $\kappa$ represents the faction alignment of an agent; $0
      \leq \kappa \leq 1$.
\end{itemize}

\subsection{Factions and Friends}

A faction is formally defined as a 3-tuple, $\Psi = \langle \gamma,
\Upsilon, \mathcal{M} \rangle$ where:
\begin{enumerate}
    \item[$\bullet$] $\gamma$ is a unique identifier for each faction.
    \item[$\bullet$] $\Upsilon$ is the set of member agents.
    \item[$\bullet$] $\mathcal{M}$, the central memory is a vector
      indexed by agent id, and each cell holds the aggregate of
      members' opinions about the corresponding agent.
    \begin{center}
    $\mathcal{M}=   \begin{bmatrix}
                    \frac{\sum_{k \in \Upsilon} \eta_k(1,t)}{|\Upsilon|} & 
                    \frac{\sum_{k \in \Upsilon} \eta_k(2,t)}{|\Upsilon|} &  
                    \hdots & 
                    \frac{\sum_{k \in \Upsilon} \eta_k(N,t)}{|\Upsilon|}\\
                    \end{bmatrix}$
    \end{center}
\end{enumerate}

Each faction is uniquely represented by an identifier and holds a set
of agents with the condition that an agent can belong to one faction
only. Each faction maintains a central memory which indicates past
levels of cooperation by all agents in the system (not just the ones
in the faction). The faction's memory is updated by all members at the
end of each interaction, and is accessible only to its members.

The number of friends are constrained as per the description in
Section~\ref{social_structure}. Agents always fully cooperate when
they interact with their friends. Any friend can access an agent's
experiences.

\subsection{Partisan Agents}

A partisan agent $\Pi$ is formally defined as a 9-tuple as given
below. A partisan agent uses all the extended attributes in addition
to the basic attributes.
\begin{center}
$\Pi = \langle \alpha,\rho,\omega,\mathcal{E},\mathcal{D},\mathcal{F},\sigma_0,\lambda,\kappa \rangle$ 
\end{center}

\noindent\emph{Behavior of a partisan agent}

Consider that agent $A$ is paired up with agent $B$ in one iteration.
The goal for agent $A$ is to come up with an optimum level of
cooperation given its own prior experiences with $B$ and the opinions
it receives from its friends.  The crux of the problem here is to come
up with the necessary Gaussian distribution, defined by $\mu, \sigma$.
Then the opinions are collected, weighed, faction value incorporated,
and then the agent cooperates at this level.  The interaction takes
place and then the agent updates the values and learns.  The process
is outlined in Algorithm~\ref{algo: partisanalgo}, and it can be
broken down into four meaningful steps as described below.

\begin{enumerate}[
leftmargin=0pt, itemindent=20pt,
labelwidth=15pt, labelsep=5pt, listparindent=0.7cm,
align=left]
    \item Initialization
    
As discussed already, the curve needs to be centered at the agent's
own opinion. So, $\mu$ is set as $A$'s opinion of $B$ formed on the
basis of its prior experiences, $\mathcal{E}_A(B)$. According to
(\ref{eq:sigmaeq}), $\sigma$ is set as product of two factors---$A$'s
base spread ($\sigma_0$) and $A$'s doubt on $B$, $\mathcal{D}_A[B]$.

Initialize two empty vectors \emph{OpinionSet} and \emph{Weights} to
capture the opinions and their respective weights. These vectors
correspond to $E$ and $W$ defined by (\ref{eq:eeq}) and (\ref{eq:weq})
respectively.  This initialization is shown in lines 1--4 of Algorithm
\ref{algo: partisanalgo}.
    
    \item Collection of opinions and assignment of weights
    
First append $A$'s opinion and its weight to \emph{OpinionSet} and
\emph{Weights} respectively. This is shown in lines 6--7, where the
function Append($l,i$) appends item $i$ to list
$l$. GaussianPDF($x,\mu,\sigma$) returns the value of Gaussian PDF
defined by $\mu$ and $\sigma$ at $x$.

Iterate through the list of $A$'s friends, and for each friend,
extract its opinion about $B$ and assign the corresponding weight
according to (\ref{eq:NormalEq}). Append the opinion and the weight to
\emph{OpinionSet} and \emph{Weights} respectively. This iteration is
captured in the for loop at lines 8--13.
    
    \item Deciding on a final level of cooperation
    
Perform a dot product on \emph{OpinionSet} and \emph{Weights} as per
(\ref{eq:doteq}) to get the intermediate decision ($O'$) based on
local opinions (Line 14). $A$ retrieves its faction's view on $B$ and
stores in \emph{FactionView}. GetFactionRating($F,X$) returns the
faction $F$'s view about an agent $B$. The final level of cooperation
is taken as the alignment-weighted average of $O'$ and
\emph{FactionView} according to (\ref{eq:finaleq}). This calculation
is shown in lines 15--17 of Algorithm~\ref{algo: partisanalgo}.
    
    \item Updating payoff and doubt
    
Calculate payoff according to (\ref{eq:payoffeq}) and update $A$'s
payoff (Line 18). Compare $B$'s level of cooperation \emph{b} with
$A$'s threshold of satisfaction and update doubt of $A$ on $B$
according to (\ref{eq:updateeq}). This condition check is done in
lines 19--23 of Algorithm \ref{algo: partisanalgo}.  The last line of
the algorithm describes the concept of sharing experiences with its
faction and that ends this interaction.
\end{enumerate}

%\vspace-0.1in}
\begin{algorithm}
    \SetKwComment{Comment}{$\triangleright$\ }{}
    \SetKwData{Op}{OpinionSet}
    \SetKwData{x}{Opinion}
    \SetKwData{Fr}{Friends}
    \SetKwData{W}{Weights}
    \SetKwData{w}{w}
    \SetKwData{mo}{$\mu$}
    \SetKwData{iexp}{$i_{exp}$}
    \SetKwData{sig}{$\sigma$}
    \SetKwData{o}{O'}
    \SetKwData{FacView}{FactionView}\SetKwData{FacAlign}{FacAlign}
    \SetKwData{LvlCoop}{LvlCoop}
    \SetKwFunction{Append}{Append}
    \SetKwFunction{dist}{GaussianPDF}
    \SetKwFunction{dot}{DotProduct}
    \SetKwFunction{GetRating}{GetFactionRating}
    \tcc*[h]{Initialize $\mu$ and $\sigma$ for Gaussian Distribution}
    
    \mo $\leftarrow$ $A$.Experience[$B$]\;
    \sig $\leftarrow$ $\sigma_0 \times D_A(B)$\;
    \Op $\leftarrow$ $\emptyset$ \;
    \W $\leftarrow$ $\emptyset$ \;
    \Fr $\leftarrow$ $A$.Friends \;

    \Append{\Op,$A$.Experience[$B$]}\;
    
    \Append{\W,\dist{$A$.Experience[$B$],\mo,\sig}}\;
    \tcc*[h]{For each friend, retrieve opinion and weight}
    
    \For{$i\leftarrow$ \Fr}{
        \iexp $\leftarrow$ i.Experience[$B$]\;
        \Append{\Op, \iexp}\;
        \w $\leftarrow$ \dist{\iexp,\mo,\sig}\;
        \Append{\W,\w}\;
    }
    \tcc*[h]{Perform dot product of \Op and \W}
    
    \o $\leftarrow$ \dot{\W,\Op} \;
    \FacView $\leftarrow$ \GetRating{$A$.factionId,$B$}\;
    \FacAlign $\leftarrow$ $A$.falign\;
    \tcc*[h]{\LvlCoop represents level of cooperation of $A$}
    
    \LvlCoop $\leftarrow$ $\o \times (1-\FacAlign) + \FacView \times \FacAlign$\;
    Calculate and update payoff\;
    \eIf {b > $A$.satisfaction} {$A$.Doubt[B] = $A$.Doubt[B] - $c$\;}
    {$A$.Doubt[B] = $A$.Doubt[B] + $c$\;}
    Share experience with faction\;
    \caption{Behavior of agent $A$ of type $\Pi$}
    \label{algo: partisanalgo}
\end{algorithm} 
%\vspace-0.2in}
\subsection{Individual Trust-Based Agents}
An individual trust-based agent $\Omega$ is defined by a 5-tuple as given below. It uses only one extended attribute, $\lambda$.
\begin{center}
    $\Omega = \langle \alpha,\rho,\omega,\mathcal{E},\lambda \rangle$
\end{center}

The only distinction here is that each cell in experiences
($\mathcal{E}$) stores the outcomes of the corresponding interaction
with that agent. The outcomes can either be 0 or 1 signifying failure
or success. We model the agents with an attribute called satisfaction
($\lambda$) to determine the outcome of an interaction. There is no
communication or sharing of experiences among these agents and they
strictly operate only based on their experiences.
%\vspace-0.1in}
\begin{algorithm}
    \SetKwData{LvlCoop}{LvlCoop}
    \SetKwFunction{Append}{Append}
    \SetKwFunction{avg}{Average}
    \tcc*[h]{$A$.Experiences[$B$] is a vector which represents previous outcomes in interactions with $B$}
    
    \LvlCoop $\leftarrow$ \avg{$A$.Experiences[$B$]}\;
    Calculate and update payoff\;
    \eIf {$b$ > $A$.satisfaction} {\Append{$A$.Experiences[$B$],1};}
    {\Append{$A$.Experiences[$B$],0};}
    \caption{Behavior of agent $A$ of type $\Omega$}
    \label{algo: trustalgo}
\end{algorithm}%\vspace-0.1in}

\emph{Behavior of an Individual trust-based agent}

Individual trust-based agents rely on their history of interactions
with other agents as their only source of information to help in
decision making. Consider a case where agent $A$ is paired with agent
$B$ in an iteration. $A$ retrieves the vector corresponding to $B$
from its Experiences vector $\mathcal{E}_A$ and calculates an average
of values and it cooperates at this level (line 1).

$A$ interacts with $B$ and payoffs are calculated (line 2) and updated
according to (\ref{eq:payoffeq}). Each agent has an attribute called
threshold of satisfaction and this helps to classify an interaction as
success or failure. If agent $B$ cooperates at a level greater than
the threshold of satisfaction ($\lambda_A$), it is classified a
success, and a failure otherwise. In case of success, the
corresponding vector is appended with 1 and in case of a failure, it
is appended with 0. This is captured in lines 3--7 of
Algorithm~\ref{algo: trustalgo}.

\subsection{Suspicious TFT Agents}

A Suspicious Tit-for-Tat (S-TFT) Agent $\Delta$ is defined by a
4-tuple and does not use any extended attribute. The only distinction
here is that their experiences vector can only capture the most recent
interaction with that agent i.e., $\omega =1$.
\begin{center}
    $\Delta = \langle \alpha, \rho,\omega,\mathcal{E} \rangle$ 
\end{center}

S-TFT agents are a standard type of agents which have been well
explored in IPD games~\cite{Axel,Boyd}.  As the name suggests, an
S-TFT agent $A$ defects completely on its first interaction with $B$
owing to its ``suspicious'' nature. However, in subsequent iterations,
$A$ cooperates at the same level that $B$ has cooperated in the
previous interaction.

\section{Experiments and Results}

\begin{figure}[!htbp]
    \setlength{\abovecaptionskip}{0pt plus 1pt minus 1pt} 
    \centering
    \includegraphics[width=0.8\textwidth, height=0.6\textheight,keepaspectratio]{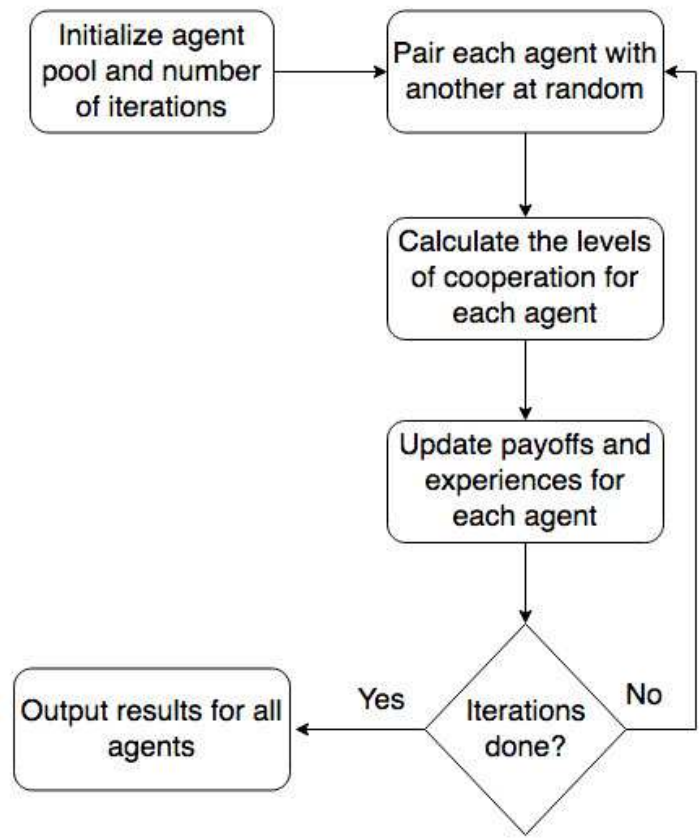}
    \caption{Workflow for system}
    \label{fig:MainFlow}
\end{figure}

The agent pool is configured with all its parameters as described in
Section~\ref{types} and in each iteration, an agent is paired randomly
with one other agent. At the end of an interaction, payoffs and
experiences are updated.  Agents capable of learning modify their
self-doubt based on the outcome. This flow is outlined by
Figure~\ref{fig:MainFlow}. We vary several parameters in the
configuration of model and individual agents' attributes such as
egocentricity to observe their effects on performance.  The findings
are presented in the following subsections.
%\vspace-0.2in}

\subsection{The Importance of Egocentricity}

\begin{figure}[!htbp]
    \setlength{\abovecaptionskip}{0pt plus 1pt minus 1pt} 
    \centering
    \includegraphics[width=0.8\textwidth, height=0.6\textheight, keepaspectratio]{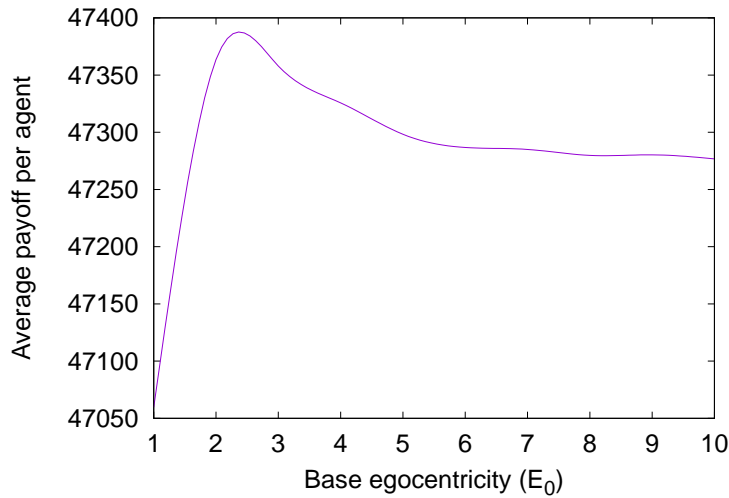}
    \caption{Effect of egocentricity on payoffs}
    \label{fig:egocentricgraph}
\end{figure}

To observe the impact of different degrees of egocentricity, we
considered a system of 500 agents equally distributed among all 3
types. We consider 5 factions in the system and vary the value of base
egocentricity ($E_0$). We find that payoffs are highest for an
intermediate level of egocentricity and is not as good for both
extremely high values and extremely low values.  Our results concur
with the conventional wisdom that egocentricity has to be at a
moderate level for better gains (Figure ~\ref{fig:egocentricgraph}).

\subsection{Comparing Payoffs of All Agent Types}

\begin{figure}[!htbp]
    \setlength{\abovecaptionskip}{0pt plus 1pt minus 1pt} 
    \centering
    \includegraphics[width=0.8\textwidth, height=0.6\textheight, keepaspectratio]{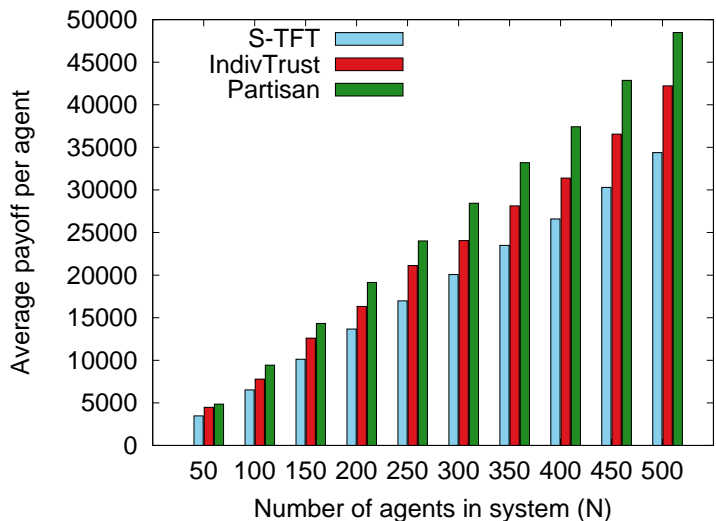}
    \caption{Comparing all types of agents}
    \label{fig:compare3}
\end{figure}

To understand the payoffs for each type and to see how they fare
against others, the system's total number of agents is varied along
with the number of factions, in such a way that each faction holds
about the same number of agents. This is done to avoid any variations
resulting from changes in faction size.  We increase the number of
agents in the system from 50 all the way up to 500.
Figure~\ref{fig:compare3} clearly indicates that partisan agents
always perform better than the other types.

\subsection{Proportion of Partisan Agents}

\begin{figure}[!htbp]
    \setlength{\abovecaptionskip}{0pt plus 1pt minus 1pt} 
    \centering
    \includegraphics[width=0.8\textwidth, height=0.6\textheight, keepaspectratio]{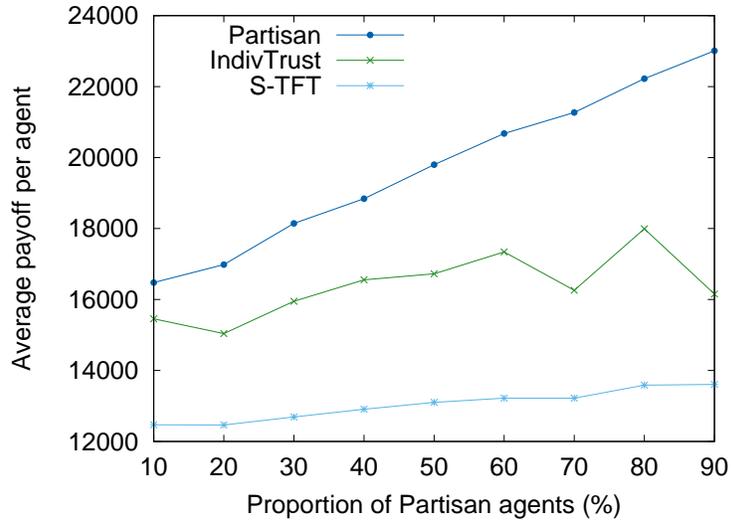}
    \caption{Effect of proportion of partisan agents on payoffs}
    \label{fig:prop}
\end{figure}

To see if partisan agents perform better at all levels of
representation in the system, we vary the proportion of partisan
agents in a system of 200 agents with 5 factions. Since we are keeping
the total number of agents and the number of factions constant and
increasing the representation of partisan agents, factions contain on
average a higher number of partisan agents, and hence their payoffs
are expected to be higher, as seen in Figure~\ref{fig:prop}.
%\vspace-0.1in}

\subsection{Effect of Faction Size}

\begin{figure}[!htbp]
    \centering
    \includegraphics[width=0.8\textwidth, height=0.6\textheight, keepaspectratio]{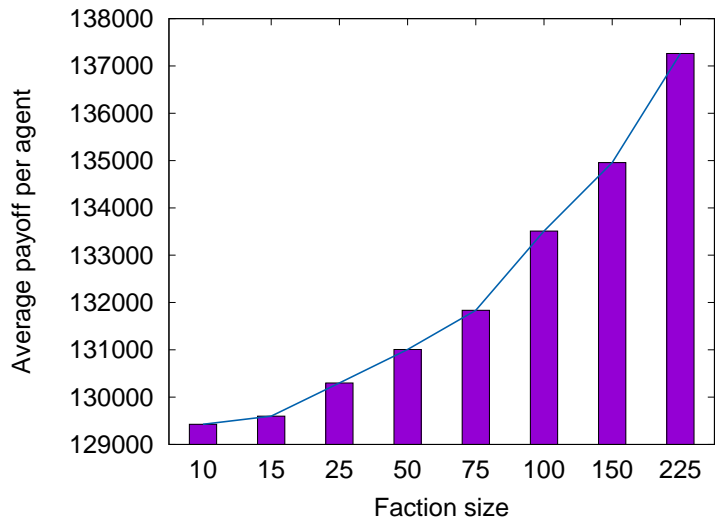}
    \caption{Effect of faction sizes on payoffs of partisan agents}
    \label{fig:facsize}
\end{figure}

To understand the effect of faction sizes on payoffs, we consider a
system of 1300 agents with half of them partisan agents and the rest
equally distributed among the other types.  We consider 10 factions in
the system with sizes ranging from 10 to 225. As the faction size
increases, the average payoff per partisan agent also increases and
this is seen in Figure ~\ref{fig:facsize}.  (Payoffs for other agent
types do not depend on faction sizes, for obvious reasons.)

\emph{Network externality} can be described as a change in the
benefit, or surplus, that an agent derives from a good when the number
of other agents consuming the same kind of good
changes~\cite{network}. Over the years, various network pioneers have
attempted to model how the growth of a network increases its
value. One such model is Sarnoff's Law which states that value is
directly proportional to size~\cite{sarnoff} (an accurate description
of broadcast networks with a few central nodes broadcasting to many
marginal nodes such as TV and radio).

Since each of our factions has one central memory that caters to all
members, it is similar to broadcast networks and
Figure~\ref{fig:facsize} exhibits a similar proportionality (with a
large offset).
%\vspace-0.2in}

\subsection{Number of Interactions and Payoffs}

\begin{figure}[!htbp]
    \setlength{\abovecaptionskip}{0pt plus 1pt minus 1pt} 
    \centering
    \includegraphics[width=0.8\textwidth, height=0.6\textheight, keepaspectratio]{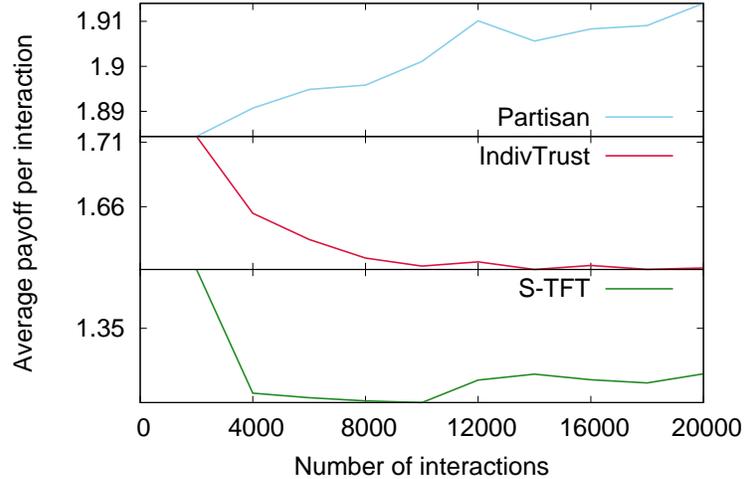}
    \caption{Number of interactions and payoffs}
    \label{fig:iters}
\end{figure}

The number of interactions is a crucial aspect when it comes to
comparing strategies because S-TFT agents may gain a lot in their
first interaction with other agents, and if there are no subsequent
interactions with the same agents, it is highly profitable for
them. However, partisan agents grow better with each interaction
because of the availability of more information.  We consider a system
of 500 agents equally distributed among all 3 types and vary the
number of interactions per agent. As expected, S-TFT agents have their
best payoffs for lower numbers of interactions, but their payoffs
start to fall rapidly with increasing interactions.  Partisan agents
steadily receive better payoffs as the number of interactions
increases (Figure ~\ref{fig:iters}).
%\vspace-0.2in}

\subsection{Number of Factions and Payoffs}

\begin{figure}[!htbp]
    \centering
    \includegraphics[width=0.8\textwidth, height=0.6\textheight, keepaspectratio]{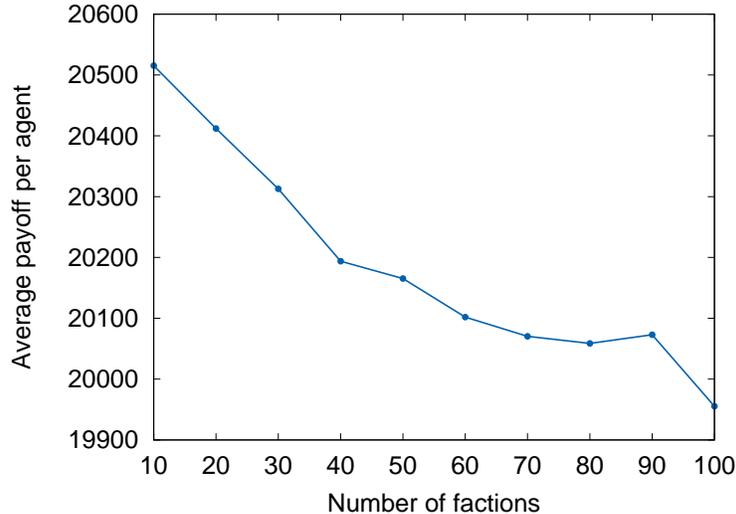}
    \caption{Number of factions and payoffs of partisan agents}
    \label{fig:numfacs}
\end{figure}

For partisan agents, the number of factions in the system plays a
vital role.  When there are many factions in the system, agents are
scattered across factions, thus weakening each faction by reducing the
information contained in the faction's central memory.  Hence, we
expect payoffs to decrease as number of factions are increased. We
have considered a system of 200 agents equally distributed among types
and vary the number of factions from 10 to 100. It is clear from the
Figure ~\ref{fig:numfacs}, that when factions are fewer in number,
partisan agents achieve high payoffs, but as the number of factions
increase, the advantage of a faction is diluted and the payoff
decreases.

%\vspace-0.1in}

\section{Conclusions}

We live in a deeply fragmented society where differences of opinion
are sometimes so high that communication may break down in some
instances. A clear model of various biases is important to understand
the underlying mechanics of how some hold opinions that may seem
irrational to others.

We present a model that closely captures the reality by imbibing the
agents with egocentric bias and doubt. We use a symmetric distribution
centered at an agent’s own opinion to assign weights to various
opinions and thus introduce more flexibility than previous models. We
also model a response to failure, by altering the self-doubt on that
topic. This balance between egocentricity and doubt enables the agent
to learn reactively.

Opinion aggregation from multiple sources is now more important than
ever owing to the effects of social media and mass
communication. Hence, there is a need for appropriate models that
realistically capture the way humans form opinions. Group opinion
dynamics continue to be an area of immense interest and hence we have
also introduced a model of a faction with a central memory. We observe
that our model of factions seems to support the theory of network
effects, and to be consistent with Sarnoff's Law.

In people, high egocentricity may be connected with anxiety or
overconfidence, and low egocentricity with depression or feelings of
low self-worth.  Our results also support the notion that
egocentricity needs to be moderate and that either extreme is not as
beneficial.

It is also observed that partisan agents generally perform much better
than the other types that have been considered, which too seems to
have parallels in human society.

\subsection*{Acknowledgements}

The author S. Rao acknowledges support from an AWS Machine Learning
Research Award.


\begin{thebibliography}{10}

\bibitem{allahverdyan}
{\sc Allahverdyan, A.~E., and Galstyan, A.}
\newblock Opinion dynamics with confirmation bias.
\newblock {\em PLOS One 9\/} (07 2014), 1--14.

\bibitem{Atwater1999}
{\sc Atwater, L.~E., Dionne, S.~D., Avolio, B., Camobreco, J.~F., and Lau,
  A.~W.}
\newblock A longitudinal study of the leadership development process:
  Individual differences predicting leader effectiveness.
\newblock {\em Human Relations 52}, 12 (1999), 1543--1562.

\bibitem{Axel}
{\sc Axelrod, R., and Hamilton, W.~D.}
\newblock The evolution of cooperation.
\newblock {\em Science 211}, 4489 (1981), 1390--1396.

\bibitem{cblist}
{\sc Benson, B.}
\newblock Cognitive bias cheat sheet, 2016.

\bibitem{herd3}
{\sc Bikhchandani, S., Hirshleifer, D., and Welch, I.}
\newblock A theory of fads, fashion, custom, and cultural change as
  informational cascades.
\newblock {\em Journal of Political Economy 100}, 5 (Oct. 1992), 992--1026.

\bibitem{selfdoubt}
{\sc Bos, K. V.~D., and Lind, E.~A.}
\newblock {\em The Social Psychology of Fairness and the Regulation of Personal
  Uncertainty}.
\newblock Routledge, 2009, ch.~7.

\bibitem{bouchet}
{\sc Bouchet, F., and Sansonnet, J.-P.}
\newblock Subjectivity and cognitive biases modeling for a realistic and
  efficient assisting conversational agent.
\newblock In {\em Proceedings of the 2009 IEEE/WIC/ACM International Joint
  Conference on Web Intelligence and Intelligent Agent Technology - Volume
  02\/} (2009), WI-IAT '09, IEEE Computer Society, pp.~209--216.

\bibitem{Boyd}
{\sc Boyd, R., and Lorberbaum, J.~P.}
\newblock No pure strategy is evolutionarily stable in the repeated prisoner's
  dilemma game.
\newblock {\em Nature 327\/} (1987), 58--59.

\bibitem{3faces}
{\sc Brown, J., and Marshall, M.}
\newblock The three faces of self-esteem.
\newblock {\em Self-esteem: Issues and answers\/} (01 2006), 4--9.

\bibitem{Burman}
{\sc Burman, O.~H., Parker, R.~M., Paul, E.~S., and Mendl, M.~T.}
\newblock Anxiety-induced cognitive bias in non-human animals.
\newblock {\em Physiology \& Behavior 98}, 3 (2009), 345 -- 350.

\bibitem{dunbar2}
{\sc Carron, P.~M., Kaski, K., and Dunbar, R.}
\newblock Calling {Dunbar}'s numbers.
\newblock {\em Social Networks 47\/} (2016), 151--155.

\bibitem{cox}
{\sc Cox, J.~T., and Griffeath, D.}
\newblock Diffusive clustering in the two dimensional voter model.
\newblock {\em Ann. Probab. 14}, 2 (04 1986), 347--370.

\bibitem{Deffuant2002}
{\sc Deffuant, G., Amblard, F., Weisbuch, G., and Faure, T.}
\newblock How can extremism prevail? a study based on the relative agreement
  interaction model.
\newblock {\em J. Artificial Societies and Social Simulation 5\/} (2002).

\bibitem{Deffuant2000}
{\sc Deffuant, G., Neau, D., Amblard, F., and Weisbuch, G.}
\newblock Mixing beliefs among interacting agents.
\newblock {\em Advances in Complex Systems 3\/} (01 2000), 87--98.

\bibitem{delVic}
{\sc Del~Vicario, M., Scala, A., Caldarelli, G., Stanley, H., and
  Quattrociocchi, W.}
\newblock Modeling confirmation bias and polarization.
\newblock {\em Scientific Reports 7\/} (2016), 40391.

\bibitem{herd6}
{\sc Dhar, J., and Jha, A.~K.}
\newblock Analyzing social media engagement and its effect on online product
  purchase decision behavior.
\newblock {\em Journal of Human Behavior in the Social Environment 24}, 7
  (2014), 791--798.

\bibitem{dunbar1}
{\sc Dunbar, R. I.~M.}
\newblock Neocortex size as a constraint on group size in primates.
\newblock {\em Journal of Human Evolution 22}, 6 (June 1992), 469--493.

\bibitem{eiser}
{\sc Eiser, J.~R., and White, M.~P.}
\newblock A psychological approach to understanding how trust is built and lost
  in the context of risk.
\newblock In {\em SCARR Conference on Trust\/} (December 2005).

\bibitem{sugarscape}
{\sc Epstein, J.~M., and Axtell, R.}
\newblock {\em Growing artificial societies: social science from the bottom
  up}.
\newblock MIT Press, 1996.

\bibitem{mem}
{\sc Fareri, D., Chang, L., and Delgado, M.}
\newblock Effects of direct social experience on trust decisions and neural
  reward circuitry.
\newblock {\em Frontiers in Neuroscience 6\/} (2012), 148.

\bibitem{Feng2018}
{\sc Feng, C., Feng, X., Wang, L., Wang, L., Gu, R., Ni, A., Deshpande, G., Li,
  Z., and Luo, Y.-J.}
\newblock The neural signatures of egocentric bias in normative
  decision-making.
\newblock {\em Brain Imaging and Behavior\/} (May 2018).

\bibitem{anchoring0}
{\sc Furnham, A., and Boo, H.~C.}
\newblock A literature review of the anchoring effect.
\newblock {\em The Journal of Socio-Economics 40}, 1 (Feb. 2011), 35--42.

\bibitem{group3}
{\sc Gill, Z.}
\newblock Wikipedia: Case study of innovation harnessing collaborative
  intelligence.
\newblock In {\em The Experimental Nature of Venture Creation: Capitalizing on
  Open Innovation 2.0}, M.~Curley and P.~Formica, Eds. Springer, Cham, 2013,
  pp.~127--138.

\bibitem{NY}
{\sc Goleman, D.}
\newblock A bias puts self at center of everything.
\newblock {\em New York Times\/} (Jun 1984).

\bibitem{Greenberg}
{\sc Greenberg, J.}
\newblock Overcoming egocentric bias in perceived fairness through
  self-awareness.
\newblock {\em Social Psychology Quarterly 46}, 2 (1983), 152--156.

\bibitem{Greenwald}
{\sc Greenwald, A.~G.}
\newblock The totalitarian ego: Fabrication and revision of personal history.
\newblock {\em American Psychologist\/} (1980), 603--618.

\bibitem{herd1}
{\sc Hamilton, W.~D.}
\newblock Geometry for the selfish herd.
\newblock {\em Journal of Theoretical Biology 31}, 2 (May 1971), 295--311.

\bibitem{rats}
{\sc Harding, E.~J., Paul, E.~S., and Mendl, M.}
\newblock Cognitive bias and affective state.
\newblock {\em Nature 427\/} (2004), 312.

\bibitem{hayashi}
{\sc Hayashi, Y., Takii, S., Nakae, R., and Ogawa, H.}
\newblock Exploring egocentric biases in human cognition: An analysis using
  multiple conversational agents.
\newblock In {\em 2012 IEEE 11th International Conference on Cognitive
  Informatics and Cognitive Computing\/} (Aug 2012), pp.~289--294.

\bibitem{Opn1}
{\sc Hegselmann, R., and Krause, U.}
\newblock Opinion dynamics and bounded confidence models, analysis and
  simulation.
\newblock {\em Journal of Artificial Societies and Social Simulation 5}, 3
  (2002), 2.

\bibitem{group2}
{\sc Hung, W.}
\newblock Team-based complex problem solving: A collective cognition
  perspective.
\newblock {\em Educational Technology Research and Development 61}, 3 (June
  2013), 365--384.

\bibitem{nowakleader}
{\sc Kacperski, K., and yst, J. A.~H.}
\newblock Opinion formation model with strong leader and external impact: a
  mean field approach.
\newblock {\em Physica A: Statistical Mechanics and its Applications 269}, 2
  (1999), 511 -- 526.

\bibitem{CPD2}
{\sc Killingback, T., and Doebeli, M.}
\newblock The continuous prisoner’s dilemma and the evolution of cooperation
  through reciprocal altruism with variable investment.
\newblock {\em The American Naturalist 160}, 4 (2002), 421--438.

\bibitem{anchoring4}
{\sc Korteling, J.~E., Brouwer, A.-M., and Toet, A.}
\newblock A neural network framework for cognitive bias.
\newblock {\em Frontiers in Psychology 9\/} (Sept. 2018), 1561.

\bibitem{sarnoff}
{\sc Kovarik, B.}
\newblock {\em Revolutions in Communication: Media History from Gutenberg to
  the Digital Age}.
\newblock Bloomsbury Publishing, 2015.

\bibitem{Krause_adiscrete}
{\sc Krause, U.}
\newblock A discrete nonlinear and non-autonomous model of consensus formation.
\newblock In {\em Communications in Difference Equations}. Gordon and Breach
  Pub., Amsterdam, 2000, pp.~227--236.

\bibitem{Krueger}
{\sc Krueger, J.~I., and Clement, R.~W.}
\newblock The truly false consensus effect: an ineradicable and egocentric bias
  in social perception.
\newblock {\em Journal of personality and social psychology 67 4\/} (1994),
  596--610.

\bibitem{kimberly}
{\sc Leister, K.~D.}
\newblock Relations among perspective taking, egocentrism, and self-esteem in
  late adolescents.
\newblock Master's thesis, University of Richmond, 1992.

\bibitem{oligopolies}
{\sc Leleno, J.~M., and Sherali, H.~D.}
\newblock A leader-follower model and analysis for a two-stage network of
  oligopolies.
\newblock {\em Annals of Operations Research 34}, 1 (Dec. 1992), 37--72.

\bibitem{lewin}
{\sc Lewin, K., Dembo, T., Festinger, L., and S.~Sears, R.}
\newblock Level of aspiration.
\newblock In {\em Personality and the behavior disorders}. 1944, pp.~333--378.

\bibitem{li2017}
{\sc Li, J., and Xiao, R.}
\newblock Agent-based modelling approach for multidimensional opinion
  polarization in collective behaviour.
\newblock {\em Journal of Artificial Societies and Social Simulation 20}, 2
  (2017), 4.

\bibitem{network}
{\sc Liebowitz, S., and Margolis, S.}
\newblock Network externality: An uncommon tragedy.
\newblock {\em Journal of Economic Perspectives 8\/} (02 1994), 133--50.

\bibitem{anchoring3}
{\sc Lieder, F., Griffiths, T.~L., Huys, Q. J.~M., and Goodman, N.~D.}
\newblock The anchoring bias reflects rational use of cognitive resources.
\newblock {\em Psychonomic Bulletin \& Review 25}, 1 (Feb. 2018), 322--349.

\bibitem{perception}
{\sc Magessi, N.~T., and Antunes, L.}
\newblock Modelling agents' perception: Issues and challenges in multi-agents
  based systems.
\newblock In {\em Progress in Artificial Intelligence\/} (Cham, 2015),
  F.~Pereira, P.~Machado, E.~Costa, and A.~Cardoso, Eds., Springer
  International Publishing, pp.~687--695.

\bibitem{nic}
{\sc Nickerson, R.~S.}
\newblock Confirmation bias: A ubiquitous phenomenon in many guises.
\newblock {\em Review of General Psychology 2}, 2 (1998), 175--220.

\bibitem{NowakLatane90}
{\sc Nowak, A., Latane, B., and Szamrej, J.}
\newblock From private attitude to public opinion: A dynamic theory of social
  impact.
\newblock {\em Psychological Review 97\/} (1990), 362--376.

\bibitem{Nowak96}
{\sc Nowak, A., and Lewenstein, M.}
\newblock Modeling social change with cellular automata.
\newblock In {\em Modelling and Simulation in the Social Sciences from the
  Philosophy of Science Point of View}. Springer Netherlands, Dordrecht, 1996,
  pp.~249--285.

\bibitem{anchoring1}
{\sc Oechssler, J., Roider, A., and Schmitz, P.~W.}
\newblock Cognitive abilities and behavioral biases.
\newblock {\em Journal of Economic Behavior and Organization 72}, 1 (Oct.
  2009), 147--152.

\bibitem{trustCB}
{\sc Pericherla, S., Rachuri, R., and Rao, S.}
\newblock Modeling confirmation bias through egoism and trust in a multi agent
  system.
\newblock The 2018 IEEE International Conference on Systems, Man, and
  Cybernetics (SMC2018).

\bibitem{Ralph}
{\sc Perry, R.~B.}
\newblock The ego-centric predicament.
\newblock {\em The Journal of Philosophy, Psychology and Scientific Methods 7},
  1 (1910), 5--14.

\bibitem{herd5}
{\sc Prechter, R.}
\newblock {\em The Wave Principle of Human Social Behavior}.
\newblock New Classics Library, 1999.

\bibitem{herd2}
{\sc Raafat, R.~M., Chater, N., and Frith, C.}
\newblock Herding in humans.
\newblock {\em Trends in Cognitive Sciences 13}, 10 (Oct. 2009), 420--428.

\bibitem{herd4}
{\sc Rook, L.}
\newblock An economic psychological approach to herd behavior.
\newblock {\em Journal of Economic Issues 40}, 1 (Mar. 2006), 75--95.

\bibitem{anchoring5}
{\sc Ross, L., Greene, D., and House, P.}
\newblock The ``false consensus effect": An egocentric bias in social
  perception and attribution processes.
\newblock {\em Journal of Experimental Social Psychology 13}, 3 (May 1977),
  279--301.

\bibitem{Ross}
{\sc Ross, M., and Sicoly, F.}
\newblock Egocentric biases in availability and attribution.
\newblock {\em Journal of Personality and Social Psychology 37\/} (1979),
  322--336.

\bibitem{Schelling}
{\sc Schelling, T.~C.}
\newblock {\em Micromotives and Macrobehavior}.
\newblock Norton, 1978.

\bibitem{esteem}
{\sc Schlenker, B.~R., Salvatore~Soraci, J., and McCarthy, B.}
\newblock Self-esteem and group performance as determinants of egocentric
  perceptions in cooperative groups.
\newblock {\em Human Relations 29}, 12 (1976), 1163--1176.

\bibitem{Silverman2007}
{\sc Silverman, B.~G., Bharathy, G., Nye, B., and Eidelson, R.~J.}
\newblock Modeling factions for ``effects based operations'': part i---leaders
  and followers.
\newblock {\em Computational and Mathematical Organization Theory 13}, 4 (Dec.
  2007), 379--406.

\bibitem{fac}
{\sc Silverman, B.~G., Bharathy, G.~K., Nye, B., , and Eidelson, R.~J.}
\newblock Modeling factions for 'effects based operations': Part i leader and
  follower behaviors.
\newblock {\em Computational and Mathematical Organization Theory 13\/} (Sep
  2007).

\bibitem{Silverman2008}
{\sc Silverman, B.~G., Normoyle, A., Kannan, P., Pater, R., Chandrasekaran, D.,
  and Bharathy, G.}
\newblock An embeddable testbed for insurgent and terrorist agent theories:
  Insurgisim.
\newblock {\em Intelligent Decision Technologies 2\/} (2008), 193--203.

\bibitem{Sobkowicz}
{\sc Sobkowicz, P.}
\newblock Opinion dynamics model based on cognitive biases.

\bibitem{group1}
{\sc Stasser, G., and Titus, W.}
\newblock Pooling of unshared information in group decision making: Biased
  information sampling during discussion.
\newblock {\em Journal of Personality and Social Psychology 48}, 6 (June 1985),
  1467--1478.

\bibitem{sznajd}
{\sc Sznajd-Weron, K., and Sznajd, J.}
\newblock Opinion evolution in closed community.
\newblock {\em International Journal of Modern Physics C 11}, 06 (2000),
  1157--1165.

\bibitem{Tamir2010}
{\sc Tamir, D.~I., and Mitchell, J.~P.}
\newblock Neural correlates of anchoring-and-adjustment during mentalizing.
\newblock {\em NAS 107}, 24 (June 2010), 10827--10832.

\bibitem{anchoring2}
{\sc Tversky, A., and Kahneman, D.}
\newblock Judgment under uncertainty: Heuristics and biases.
\newblock {\em Science 185}, 4157 (Sept. 1974), 1124--1131.

\bibitem{CPD}
{\sc Verhoeff, T.}
\newblock {\em A continuous version of the prisoner's dilemma}.
\newblock Computing science notes. Technische Universiteit Eindhoven, 1993.

\bibitem{Weibush}
{\sc Weisbuch, G., Deffuant, G., Amblard, F., and Nadal, J.-P.}
\newblock Interacting agents and continuous opinions dynamics.
\newblock In {\em Heterogenous Agents, Interactions and Economic Performance\/}
  (Berlin, Heidelberg, 2003), R.~Cowan and N.~Jonard, Eds., Springer Berlin
  Heidelberg, pp.~225--242.

\bibitem{woodman}
{\sc Woodman, T., Akehurst, S., Hardy, L., and Beattie, S.}
\newblock Self-confidence and performance: A little self-doubt helps.
\newblock {\em Psychology of Sport and Exercise 11}, 6 (2010), 467--470.

\end{thebibliography}
\end{document}